\documentclass{article}

\usepackage{microtype}
\usepackage{graphicx}
\usepackage{subfigure}
\usepackage{booktabs} 

\usepackage[algo2e]{algorithm2e}

\usepackage[accepted]{icml2019}

\usepackage{amsthm}
\usepackage{amsmath}
\usepackage{amssymb, bm}
\usepackage{mathrsfs} 
\usepackage{mathtools}
\usepackage[capitalize]{cleveref}

\def\[#1\]{\begin{align}\centering#1\end{align}}
\newcommand{\defas}{\vcentcolon=}  
\newcommand{\given}{\mid}

\newcommand{\Reals}{\mathbb{R}}

\def\EE{\mathbb{E}}

\newcommand{\Normal}{\mathcal{N}}
\newcommand{\dist}{\ \sim\ }

\newcommand{\Bernoulli}{\mathrm{Bernoulli}}
\newcommand{\betadist}{\mathrm{beta}}

\newcommand{\Multinomial}{\mathrm{Mult}}

\newcommand{\Lcal}{\mathcal{L}}

 \newcommand{\params}{\theta}
 \newcommand{\nnet}{f_\params}

  \newcommand{\lnorm}[1]{\vert\vert #1 \vert\vert_2}
 \newcommand{\fnorm}[1]{\vert\vert #1 \vert\vert_\text{F}}
 
 \newcommand{\nnetreg}{\lambda}
 \newcommand{\obs}{\mathcal O}

\icmltitlerunning{Variational neural network matrix factorization and stochastic block models}

\begin{document}

\twocolumn[
\icmltitle{Variational inference for neural network matrix factorization\\
and its application to stochastic blockmodeling}

\icmlsetsymbol{equal}{*}

\begin{icmlauthorlist}
\icmlauthor{Onno Kampman}{gs}
\icmlauthor{Creighton Heaukulani}{un}
\end{icmlauthorlist}

\icmlaffiliation{gs}{Goldman Sachs, Hong Kong}
\icmlaffiliation{un}{No affiliation}

\icmlcorrespondingauthor{Onno Kampman}{onno.kampman@gs.com}
\icmlcorrespondingauthor{Creighton Heaukulani}{c.k.heaukulani@gmail.com}

\icmlkeywords{Bayesian inference, network models, collaborative filtering, relational data, deep learning, neural network}

\vskip 0.3in
]


\printAffiliationsAndNotice{}

\begin{abstract}
We consider the probabilistic analogue to neural network matrix factorization (Dziugaite \& Roy, 2015), which we construct with Bayesian neural networks and fit with variational inference. We find that a linear model fit with variational inference can attain equivalent predictive performance to the regular neural network variants on the Movielens data sets.
We discuss the implications of this result, which include some suggestions on the pros and cons of using the neural network construction, as well as the variational approach to inference.
Such a probabilistic approach is required, however, when considering the important class of stochastic block models. We describe a variational inference algorithm for a neural network matrix factorization model with nonparametric block structure and evaluate its performance on the NIPS co-authorship data set.
\end{abstract}

\section{Introduction}

Matrix factorization models are an important class of machine learning methods, playing a prominent role in dimensionality reduction, with applications to product recommendations in commerce, among others.
For example, $X_{n,m}$ could represent the amount of item $m\le M$ purchased by user $n\le N$. A classic approach to factorizing the $N\times M$ matrix $X$ would assume a linear model such as
\[
X_{n,m} = U_n^T V_m = \sum_{k=1}^K U_{n,k} V_{m,k}
    ,
    \:\: n\le N, m\le M,
    \label{eq:linear-mf}
\]
for some (relatively small) number of factors $K\ll N,M$, and where the parameter vectors $U_n$ and $V_m$ are to be inferred with a procedure such as singular value decomposition.
\citet{dziugaite2015neural} consider a \emph{neural network matrix factorization} alternative that 
replaces the linear function in \cref{eq:linear-mf} with a feed-forward neural network (with inputs $U_n$ and $V_m$), which improves predictive performance when predicting missing entries of the matrix.

Here we take a probabilistic approach by using a \emph{Bayesian neural network}, and we fit the parameters of the model with variational inference. While \emph{probabilistic matrix factorization} \citep{mnih2008probabilistic} has shown improvements (for linear models) over its non-Bayesian counterpart, we find only a small improvement for this Bayesian variant of neural network matrix factorization (fit via variational inference, anyway) upon the predictive performance of the neural network on the Movielens 100K and 1M data sets. However, we do find that variational inference can get a linear model to match the performance of the neural network, and that the neural network structure provides further improvements when side information (such as the genre of the film) is included.
In light of this (rather surprising) result, we provide a discussion on the pros and cons of using neural network structures and/or variational inference in these contexts.

Finally, one case when a probabilistic approach is \emph{required} for tractable inference is in the important class of stochastic block models. We present a variant of neural network matrix factorization applied to network models (i.e., the matrix $X$ is symmetric in this case) that captures nonparametric block structure, similar in spirit to the \emph{infinite relational model} \citep{kemp2006learning}. We derive the variational inference procedure for such a model, and we show that its predictive performance improves upon its linear analogues when applied to the NIPS co-authorship data set.

\section{Neural network matrix factorization}

%
Following \citet{dziugaite2015neural}, model the entries of $X$ as the outputs from a neural network $\nnet$ with parameters $\params$, whose inputs are (unobserved) \emph{features} of the users and items. In particular, for every $n\le N$ and $m\le M$, let
\[
X_{n,m} = \nnet([U_n, V_m, U'_{n,1} \circ V'_{m,1}, \dotsc, U'_{n,D} \circ V'_{m,D}])
    ,
    \nonumber
\]
where the parameters have the following shapes: $U_n, V_m \in \Reals^K$, and $U'_{n, d}, V'_{m,d} \in \Reals^{K'}$, $d\le D$, for some selected $K$, $K'$, and $D$.
The notation $\circ$ here denotes the element-wise product, and $[a, b, \dotsc]$ denotes the \emph{vectorization} function, i.e., the vectors $a$, $b$, $\dotsc$ are concatenated into a single vector.
Note that this neural network has $2K + K'D$ inputs and a univariate output.

Classic, linear constructions of the matrix factorization model can be recovered by restricting $\nnet$ to be a linear function.
The vectors $U'_{n,d} \circ V'_{m,d}$ play an analogous role to the traditional \emph{bilinear} terms in the linear variants of matrix factorization, and the terms $U_n$ and $V_m$ play the role of the user- and item-specific \emph{bias} terms in modeling variants such as those presented by \citet{koren2009matrix}.

Inference in this model could then minimize the following regularized squared error loss function
\[
&\sum_{(n,m) \in \obs}
    (X_{n,m} - \hat X_{n,m})^2
    +
    \nnetreg\, \cdotp \Bigl [
        \sum_n \lnorm{U_n}^2  
        \nonumber
        \\
        &\qquad \quad
        + \sum_m \lnorm{V_m}^2
        + \sum_n \fnorm{U'_n}^2
        + \sum_m \fnorm{V'_m}^2
    \Bigr ]
    ,
    \\
    &\hat X_{n,m} 
    = 
    \nnet([U_n, V_m, U'_{n,1} \circ V'_{m,1}, \dotsc, U'_{n,D} \circ V'_{m,D}])
    ,
    \nonumber
\]
where $\obs$ denotes the set of observed edges, $\fnorm{A}$ denotes the Frobenius norm for an array $A$, and $\nnetreg >0$ is a regularization parameter.

\section{Stochastic variational inference}
\label{sec:vi}

We consider letting $\nnet$ be a Bayesian neural network and elect a mean-field variational approach to inference.
In the Bayesian perspective, the likelihood of the parameters given the data is conditionally Gaussian
\begin{gather}
X_{n,m} \given \mu_{n,m} \dist \Normal(\mu_{n,m}, \sigma^2)
    \\
    \mu_{n,m} = \nnet([U_n, V_m, U'_{n,1} \circ V'_{m,1}, \dotsc, U'_{n,D} \circ V'_{m,D}])
    ,
    \nonumber
\end{gather}
for every $n\le N$, $m\le M$ and some additional noise parameter $\sigma>0$.
The components of the input arrays $U$, $V$, $U'$, and $V'$ are all given independent mean zero Gaussian prior distributions (with array-specific, shared variance parameters), as are the weights and biases in $\params$.

We follow \citet{salimans2013fixed, kingma2013auto} to implement a gradient-based variational inference routine, where minibatches are subsampled from the observed edges in the graph, and where the required gradients are estimated by low-variance Monte-Carlo approximation routines. This technique is applied to both the neural network parameters $\params$ and the inputs $U, V, U', V'$, which are updated in alternating steps during the gradient descent routine.
This has become a common practice for variational inference with Bayesian neural networks, and so we defer the reader to the references for technical details.

\subsection{Exploration of the linear model}
\label{sec:movielens}

\begin{table*}[t]
\caption{RMSE scores for the Movielens data sets. The results for Bias-MF, NN(3), and NN(4) are taken from \citet{dziugaite2015neural}.}
\label{tab:movielens}
\vskip 0.15in
\begin{center}
\begin{small}
\begin{sc}
\begin{tabular}{lcccc|cccc}
\toprule
Data set & SVD & Bias-MF & NN(3) & NN(4) & VI(0) & VI(3) & VI(0)+S & VI(3)+S\\
\midrule
Movielens 100K & 0.987 & 0.911 & 0.907 & 0.903 & 0.903 & 0.902 & 0.900 & 0.898 \\
Movielens 1M & 0.917 & 0.852 & 0.846 & 0.843 & 0.839 & 0.836 & - & - \\
\bottomrule
\end{tabular}
\end{sc}
\end{small}
\end{center}
\vskip -0.1in
\end{table*}

We ran experiments on the Movielens 100K and Movielens 1M data sets~\citep{harper2016movielens}, which contain $N=943$ users and $M=1,682$ items (with 100,000 observations) and $N=6,040$ users and $M=3,706$ items (with 1,000,209 observations), respectively. Following the experimental setup of \citet{dziugaite2015neural}, we create five random training/testing splits of the data sets, where 10\% of the data set is held out as a test set in each instance. The root mean squared error (RMSE) is displayed for various models in \cref{tab:movielens}.

The results from \citet{dziugaite2015neural} using a neural network for $\nnet$ with hidden layers, each with 50 sigmoidal units, are reported as NN(3) and NN(4), and the models fit with variational inference as VI(0) and VI(3). In all of these variants, $K=10$, $D'=60$, and $K'=1$. The VI models adapted the learning rates using Adam \citep{kingma2014adam}, with an initial learning rate of 0.001.
Batch learning (i.e., no minibatches) was used for all models. Due to memory constraints, we used training minibatches of 30,000 for the 1M data set.
For reference, we have also included a singular value decomposition (SVD) baseline (truncated at 60 singular values), and the \emph{biased matrix factorization} (Bias-MF) model \citep{koren2009matrix}.

Rather surprisingly, with variational inference we were able to get a linear model to match the performance of the neural network architecture. 
One possible conclusion is that variational inference is simply better at model selection than even a fine grid search.
A Bayesian neural network fit with mean-field variational inference has the interpretation of placing a separate L2 regularization parameter (associated with the variance parameters of the Gaussian distributed components of the variational distribution) on each weight (and possibly bias) parameter of the function $\nnet$. This is rarely done in the non-Bayesian approaches to training neural neural networks, where typically a single or very few such regularization parameters are shared across the weights of the network.
Moreover, with variational inference, these (possibly very many) weight regularization parameters are fit during gradient descent, whereas in non-Bayesian approaches they are typically selected by grid searching across multiple inference runs, which are easy to implement in parallel with the appropriate computing infrastructure, though can be a bit cumbersome to do so systematically.
We note that \citet{dziugaite2015neural} did not regularize the parameters of $\nnet$ in their experiments. However, it's still a useful (if unsurprising) lesson to see that within a single run of the inference procedure, variational inference is able to seamlessly do an otherwise piecemeal computational task. There is a slightly larger computational burden associated with variational inference, however, since the number of parameters to fit during inference doubles. Computations also increase linearly with the number of Monte Carlo samples used to approximate the required gradients (see \citet{salimans2013fixed} and \citet{kingma2013auto}), though this number can usually be very small (often one).

Viewed alternatively, the performance of the neural network suggests that by using its expressive power along with modern techniques in gradient-based inference, a user may largely ignore careful model selection on the weights of the neural network, or exhaustively fine grid searches over the regularization parameter $\nnetreg$.

\subsection{Incorporating side information}

For the Movielens 100K experiments, we also included the genre of each film as side information into the model, concatenated to the movie embedding $V_m$ in the form of a one-hot vector. There are 19 different genres. The results are presented in \cref{tab:movielens} as VI(0)+S for the linear model and VI(3)+S for a neural network with 3 hidden layers of 50 units each. We can see that the performance of both models improves, perhaps suggesting that the nonlinear structure of the neural network is advantageous when handling (observed) side information. 

\section{Stochastic block models for network data}

In this section, we will restrict our attention to the special case of network data sets, where the rows and columns of an $N\times N$ data matrix $X$ correspond to the same set of $N$ users, and an entry $X_{i,j} = 1$ if there is a ``link'' between users $i$ and $j$ and $X_{i,j}=0$ otherwise.
Such models are appropriate for social networks, where links represent friendships between individuals. 
We further assume the matrix $X$ is symmetric (i.e., $X_{i,j}=X_{j,i}$), and we do not allow self-links (i.e., the diagonal elements of $X$ are meaningless). 

In the previous section, we considered some pros and cons of optionally using a Bayesian neural network $\nnet$. However, one scenario where a Bayesian approach is \emph{required} for tractable inference is with \emph{stochastic block models} \citep{kemp2006learning, airoldi2008mixed}. In this important class of ``community detection'' models for network data, the users are clustered into groups, and the parameters of the model are shared amongst the members of a group in order to capture a well-observed phenomenon known as \emph{homogeneity}. For example, clusters in a social network could represent shared interests of the users, or geographic location, both of which presumably increase the likelihood that those users will be linked. 

We take a nonparametric, Bayesian approach to stochastic blockmodeling, in a similar spirit to the \emph{infinite relational model} by \citet{kemp2006learning}, which uses the Dirichlet process to model a potentially unbounded number of clusters that is inferred from the data.
For every $i\le N$, let $Z_i \in \{1, 2, \dotsc\}$ denote the (random) assignment of user $i$ to one of an unbounded number of groups. For every $c = 1, 2, \dotsc$, let $U_c \in \Reals^K$ and $U'_{c,d} \in \Reals^{K'}$, $d\le D$, denote the shared input features for the users in group $c$.
Then for every $i<j\le N$, let
\begin{gather}
X_{i,j} \given p_{i,j} \dist \Bernoulli( p_{i,j} )
    \\
    p_{i,j} = \nnet([U_{Z_i}, U_{Z_j}, U'_{Z_i,1} \circ U'_{Z_j,1}, \dotsc, U'_{Z_i,D} \circ U'_{Z_j,D}])
    ,
    \nonumber
\end{gather}
where the neural network $\nnet$ is now specified so that its output layer is pushed through a mapping to $(0,1)$, such as the logistic sigmoid function.

The distribution on the assignment variables $Z\defas (Z_1, \dotsc, Z_n)$ is given by the (assignments under a) \emph{Dirichlet process mixture model}, which we may describe via the \emph{stick-breaking construction} for the Dirichlet process \citep{sethuraman1994constructive}.
Independently for every $i\le N$, let $Z_i \given \pi \dist \pi$ be a sample in $\{1,2,\dotsc\}$ according to the (infinite dimensional) probability vector $\pi\defas (\pi_1, \pi_2, \dotsc)$ defined as follows
\begin{gather}
\pi_i = V_i \prod_{j=1}^{i-1} (1-V_j)
    ,
    \quad i = 1, 2, \dotsc ,
    \\
    V_i \dist \betadist(1, \alpha)
    ,
    \quad i=1, 2, \dotsc 
    ,
\end{gather}
where $\sum_{i=1}^\infty \pi_i = 1$, almost surely (as required), and $p(Z_n \given \pi) = \pi_{Z_n}$, for every $n\le N$, and $\alpha>0$ is some \emph{concentration parameter}.

The likelihood of the parameters given the data is then
\[
\Lcal
&=
p(\params) \prod_{i=1}^\infty \betadist(V_i; 1, \alpha)
    \prod_{n=1}^N \Bigl [ p(Z_n\given \pi) p(U_n) p(U'_n) \Bigr ]
    \nonumber
    \\
    &\qquad \times
    \prod_{i<j\le N} \Bernoulli(X_{i,j}; p_{i,j}) 
    ,
\]
where $p(U_n)$, $p(U'_n)$, and $p(\params)$ are the usual component-wise Gaussian densities for the inputs and neural network parameters specified in \cref{sec:vi}.

\begin{table}[t]
\caption{RMSE and AUC scores on the NIPS co-authorship data set. Bias-MF is non-Bayesian. VI and SBM are linear, since additional layers did not improve results. Note that SBM has significantly fewer parameters than the other models.}
\label{tab:sbm-results}
\vskip 0.15in
\begin{center}
\begin{small}
\begin{sc}
\begin{tabular}{lccccccc}
\toprule
Metric  & SVD & Bias-MF & VI & SBM \\
\midrule
RMSE  & 0.136 & 0.125 & 0.120 & 0.128 \\
AUC  & 0.707 & 0.839 & 0.844 & 0.718 \\
\bottomrule
\end{tabular}
\end{sc}
\end{small}
\end{center}
\vskip -0.1in
\end{table}
%
%

\subsection{Gradient-based variational inference}

We follow \citet{blei2006variational} and take a mean-field variational approach to inference with this model, in which the discrete variables $Z$ are integrated out, turning an intractable inference task into an optimization of continuous variables. The number of groups is also automatically inferred during this process. In particular, the variational approximation introduces a \emph{truncation level} as the maximum number of components of the Dirichlet process. In practice, this truncation is selected to be large enough so that the algorithm does not ``exhaust'' all available components.
Let
\[
q(Z, V) 
    = \prod_{i=1}^N \Multinomial(Z_i; \eta_i) 
    \prod_{c=1}^T \betadist(V_c ; \rho_{c,1}, \rho_{c,2})
\]
denote the mean-field variational approximation, where $\eta_i$ is a $T$-dimensional probability vector for some selected truncation level $T$, and $\rho_{c,1}, \rho_{c,2}>0$.

The parameters $\eta_i$ may be updated analytically following derivations similar to those by \citet{blei2006variational} as follows. For every $i\le N$ and $t\le T$,
\[
\label{eq:update-eta}
\eta_{i,t}
    \propto 
    \exp &\Bigl \{ 
        \EE_q [ \log V_t ] + \sum_{\ell=1}^{t-1} \EE_q [ \log (1-V_\ell) ]
        \\
        &
        + \sum_{j \colon (i,j) \in \obs} \EE_q [ \log \Bernoulli(X_{i,j} \given p_{i,j}) ]
    \Bigr \}
    ,
    \nonumber
\]
where
$
\EE_q [ \log V_t ] = \psi( \rho_{t,1} ) - \psi( \rho_{t,1} + \rho_{t,2} )
$
and
$
\EE_q [ \log (1-V_t) ] = \psi( \rho_{t,2} ) - \psi( \rho_{t,1} + \rho_{t,2} )
    ,
$
with $\psi(a) \defas \Gamma'(a)/\Gamma(a)$ denoting the digamma function, and where the term $\EE_q [ \log \Bernoulli(X_{i,j} \given p_{i,j}) ]$ is approximated with a Monte-Carlo estimate.

The variational parameters $\rho_{c,1}$, $\rho_{c,2}$ also have analytic updates, however, we found it more successful to infer them with gradient-based updates. The concentration parameter $\alpha$ is optimized directly with gradient-based updates (i.e., type-I maximum likelihood). Finally, the inputs $U$ and $U'$ and the neural network parameters $\params$ are inferred in the usual way (specified in \cref{sec:vi}).
The parameter update schedule we followed is shown in \cref{alg:sbm-svi}.
\begin{algorithm}[t]
 \KwData{$N\times M$ matrix $X$.}
 \Repeat{Convergence}{
  \nl Sample a minibatch of the edges $\obs^b \subset \obs$.\\
  \nl For every node $n$ present in (an edge in) the minibatch $\obs^b$, update $\eta_n$ according to \cref{eq:update-eta} with gradients approximated on $\obs^b$.\\
  \nl Update $q(V)$ and $\alpha$.\\
  \nl Update $q(\params)$ with gradients approximated on $\obs^b$.\\
  \nl For every node $n$ present in the minibatch $\obs^b$, update $q(U_n)$ and $q(U_n')$ with gradients approximated on $\obs^b$.
 }
 \caption{Stochastic variational inference for the stochastic block model}
 \label{alg:sbm-svi}
\end{algorithm}

\subsection{Exploring the NIPS co-authorship dataset}
\label{sec:nips}

We ran experiments on the NIPS 1--17 co-authorship data set \citep{nipsdata}, consisting of authors that had published at least nine papers at NIPS between 1988 and 2003 (resulting in $N=234$ authors). A link occurs between two authors if they co-authored at least one paper.
A truncation level of $T=7$ was used in the variational approximation to the Dirichlet process, and we note that these did not appear to be ``exhausted'' in our experiments.
The experimental setup (five randomly held out test sets) and hyperparameter settings are otherwise identical to those in \cref{sec:movielens}. The RMSE and AUC scores (averaged over the training runs and test sets) are reported in \cref{tab:sbm-results}. Note that the (non-Bayesian) neural network matrix factorization model with no hidden layers is equivalent to the biased matrix factorization model ``Bias-MF'', and so we use that name here. Bias-MF and its Bayesian analogue (fit with variational inference) ``VI'' only slightly best the linear variant of the stochastic block model ``SBM'', which is remarkable since the stochastic block model has significantly fewer features. In particular, note SBM uses $T*(K+K'*D)$ input parameters, whereas Bias-MF and VI use $N*(K+K'*D)$. This difference is perhaps more pronounced, since the properties of the Dirichlet process attempt to effectively ``pinch out'' some of these features. 
Additional layers did not improve results here.

\section{Future directions}

On one hand, our results suggest investigation into models constructed from neural networks on whether their success depends on increasing model capacity/complexity. On the other hand, conventional wisdom has always suggested that more parsimonious models generalize better to new data, though that does not seem to be a hindrance to the neural network models in our experiments.
Finally, the apparent advantages of the neural network when incorporating side information should be further explored.

\bibliography{final}
\bibliographystyle{icml2019}

\end{document}